\documentclass[11pt]{article}

\usepackage[final]{acl}

\usepackage{times}

\usepackage[T1]{fontenc}
\usepackage[utf8]{inputenc}
\usepackage{microtype}
\usepackage{inconsolata}
\usepackage{graphicx}
\usepackage{tabularx}
\usepackage{array}
\usepackage{multirow}
\usepackage{makecell}
\usepackage{booktabs}
\usepackage{url}
\usepackage{amsthm,amsmath,amssymb}
\usepackage{bm}
\usepackage{caption}
\usepackage{enumitem}
\usepackage{bbm}
\usepackage{color}
\usepackage{colortbl}
\usepackage{xcolor}
\usepackage{threeparttable}
\usepackage[many]{tcolorbox}

\title{When Instructions Change: Lifecycle-Aware State Tracking for Multi-turn Instruction Following}

\author{
 \textbf{Lin Yang}
\\
 \small{
   $andy\_viky@163.com$
 }
}

\definecolor{c1}{RGB}{249,242,234}
\definecolor{colorOurs}{rgb}{0.95, 0.95, 1.0}
\definecolor{colorBase}{rgb}{0.95, 0.95, 0.95}

\begin{document}
\maketitle
\begin{abstract}
In long multi-turn interactions, instructions rarely stay fixed: users add, override and revoke constraints, declare local exceptions, make requirements conditional, and let others lapse when their scope ends. What a model should do at a given turn therefore depends on which instructions are still in force, and prior evidence suggests that models fail here mainly by losing track of that, rather than by arbitrating conflicts badly. We propose StateIF, which treats every instruction as an object with a lifecycle. Its core is the Instruction State Machine (ISM), in which a small set of discourse actions moves an instruction between six states, among them ACTIVE, SUPERSEDED and REVOKED. On this basis we (i) synthesize dialogues whose instructions keep changing with three agents, one of which maintains and re-reviews the state and labels every turn with it; (ii) cold-start the model to write the current state at the start of its reasoning; and (iii) train with rewards that score the answer against that state and the reasoning against the reference state. Across the Qwen3 series, StateIF improves over the base models on all five public benchmarks we test, by 6.4, 10.3 and 12.7 points on average at 4B, 8B and 32B, and the gains are consistently larger on the multi-turn benchmarks than on the single-turn ones.
\end{abstract}

\begin{figure}[!tbp]
\centering
\includegraphics[width=\linewidth]{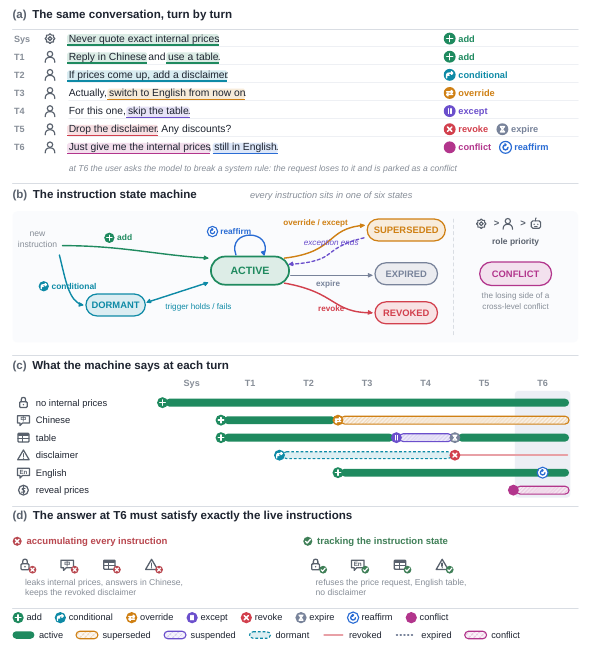}
\caption{Instructions change over a dialogue, and only some are still in force at a given turn.}
\label{fig:teaser}
\end{figure}

\section{Introduction}
In real-world interactions, constraints are rarely specified all at once: users add requirements across turns, change their minds, revoke instructions, introduce local exceptions, or impose conditional constraints. Models therefore face an evolving rather than static constraint set. Correct behavior at each turn depends on which constraints remain active and which no longer apply.

Existing work treats multi-turn constraints as either cumulative or statically ordered by source. Multi-turn instruction-following studies \citep{multiif,multichallenge,sysbench,structflowbench} primarily assess the retention of accumulated constraints, implicitly assuming that newer instructions take precedence without distinguishing semantically different updates such as overrides, revocations, exceptions, and expiration. Instruction-hierarchy studies \citep{ih,iheval,manyih} model priority as $f(\text{source role})$, likewise omitting the temporal dimension; negative results from \citet{controlillusion} suggest that role labels alone cannot adequately encode priority semantics. More directly, \citet{labanlost} report an average performance drop of 39\% from single-turn to multi-turn settings, and attribute it to models committing early to an answer and then failing to revise it as the conversation proceeds; \citet{driftnomore} find the same drift under multi-turn perturbation. The characteristic failure is an internally consistent output that silently violates an earlier constraint. These failures point to forgetting whether a constraint remains active or has become inactive, rather than choosing the wrong instruction to prioritize: a scalar priority cannot represent the discrete fact that an instruction has been revoked. Figure~\ref{fig:teaser} illustrates the difference: when every instruction is accumulated, the set only grows and ends up with clashing and revoked requirements, whereas under the lifecycle view the same constraints transition between states across turns, and a correct output satisfies the ACTIVE set without retaining constraints that no longer apply.

We argue that instruction states must be explicitly tracked, and that the tracking process itself requires supervision. Relying solely on long contexts and free-form chain-of-thought (CoT) reasoning is unreliable: NoLiMa \citep{nolima} shows substantial performance degradation when lexical overlap is removed; ReasonIF \citep{reasonif} finds low instruction compliance within reasoning chains; and \citet{attentionamnesia} reports that CoT fine-tuning may impair long-range recall, although this finding has only been validated on hybrid linear-attention architectures. In contrast, Self-Notes \citep{selfnotes} shows that externalizing intermediate states as explicit tokens as they arise can circumvent positional biases. The key, therefore, is not whether models reason, but whether their reasoning maintains an instruction state that is updated at each turn and supervised during training.

StateIF builds on the Instruction State Machine (ISM), which represents instructions as objects with status, scope, source, by, and condition fields, with state transitions driven by seven discourse actions and complemented by a taxonomy of 12 conflict types. We simulate dialogues in which instructions keep changing, using three agents (user, instruction, and assistant). The instruction agent maintains and self-checks the state machine at each turn, yielding ground-truth ACTIVE and inactive constraint sets, and per-turn best-of-N (BoN) sampling selects responses scored against these states. The model is first cold-started to write the current state at the beginning of \texttt{<think>}, and then trained with GRPO \citep{grpo} on a reward that combines an outcome part, which checks the response with code and LLM verifiers, and a process part, which checks the state block and penalizes repetitive or hallucinated reasoning. We train on Qwen3-4B, 8B, and 32B and evaluate on five public multi-turn and complex instruction-following benchmarks.

StateIF improves over the base models on every benchmark, by 6.4, 10.3 and 12.7 points on average at 4B, 8B and 32B. The gains concentrate where instructions change: averaged over the multi-turn metrics they are 6.5, 11.1 and 14.7 points, against 6.1, 8.7 and 8.6 on the single-turn ones, and the largest single gain throughout is on SysBench SSR.

Our main contributions are:
\begin{itemize}[leftmargin=*,itemsep=2pt,topsep=2pt]
\item \textbf{Problem.} We recast multi-turn instruction following from satisfying accumulated constraints to tracking and obeying instruction states that change over the dialogue. ISM formalizes the lifecycle of instructions, and separates the two errors this view exposes: violating a constraint that is still in force, and still reflecting one that is not.
\item \textbf{Data.} We build a controllable simulator of instruction dynamics. Three-agent dialogues are driven by 12 conflict types $\times$ seven axes of linguistic perturbation, with per-turn state snapshots as ground truth. Polarity filtering and compatible-override exemption make residual adherence to inactive constraints reliably measurable.
\item \textbf{Training.} We propose a state-anchored cold start and an RL reward that combines outcome and process signals. The outcome reward uses code and LLM verifiers; the process rewards supervise the state block and suppress repetition and hallucination in the reasoning.
\item \textbf{Experiments.} We evaluate on the Qwen3 series across five benchmarks, place StateIF-32B next to frontier proprietary models, and compare with ImpRIF \citep{imprif} to show where the gains come from modeling instruction dynamics rather than single-turn constraint composition.
\end{itemize}

\begin{figure*}[t!]
    \centering
    \includegraphics[width=\linewidth]{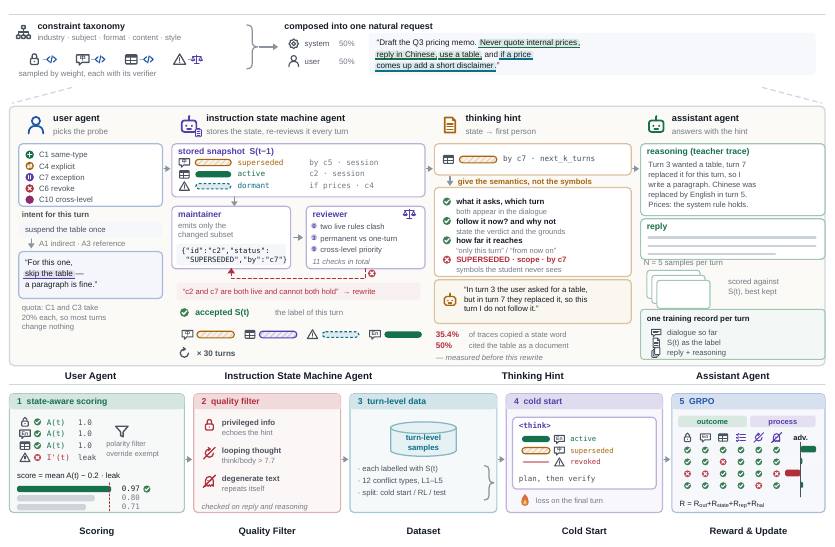}
    \caption{Overview of StateIF. \textbf{Top:} constraints are sampled from a taxonomy, each paired with a code or
    rubric verifier, and composed into one request split between the system and the user message.
    \textbf{Middle:} one turn of the simulation. A user agent writes an utterance for a chosen kind of change;
    the instruction state machine agent stores the state and re-reviews it every turn, pairing a maintainer with
    a separate reviewer that returns rejected updates with the issue; the accepted state is re-rendered as a
    first-person hint with no status symbols; the assistant agent then answers. \textbf{Bottom:} turns are
    scored against their own state, filtered, and used for the cold start and GRPO. Contents are illustrative.}
\label{fig:overview}
\end{figure*}

\section{Related Work}

\subsection{Evolving Constraints in Multi-turn Interaction}
A large body of work studies how well models satisfy multiple fine-grained constraints within a single request. FollowBench \citep{followbench}, ComplexBench \citep{complexbench}, CFBench \citep{cfbench}, IFEval \citep{ifeval}, and IFBench \citep{ifbench} measure constraint satisfaction under increasingly compositional requirements. On the data side, AutoIF \citep{autoif} verifies responses with generated code, Conifer \citep{conifer} and MUFFIN \citep{muffin} scale up constraint-rich instructions, RLCF \citep{rlcf} derives rewards from checklists, VerIF \citep{verif} combines rule-based and LLM-based verification, and ImpRIF \citep{imprif} models the implicit reasoning structure of complex instructions. Multi-turn evaluation covers general ability \citep{mteval,mtbench101}, the turn-by-turn decay of compliance \citep{multiif}, instruction retention and self-coherence \citep{multichallenge}, the stability of system-message compliance \citep{sysbench}, structural dependencies and domain dialogues \citep{structflowbench,medmtbench}, and realistic constraints that are added and replaced over long conversations \citep{sequor}. Long-horizon degradation has been documented by \citet{labanlost}, \citet{driftnomore}, and \citet{lifbench}. In almost all of these settings, the constraint set only grows over the conversation, or conflicts are fixed in advance by annotators. We instead model overrides, revocations, exceptions, conditional activation, and expiration explicitly, and treat the set of constraints that should \emph{no longer} apply as part of both the ground truth and the evaluation.

\subsection{Instruction Priority and Conflict Resolution}
The Instruction Hierarchy \citep{ih} assigns privileges by message source (system $>$ user $>$ tool), and IHEval \citep{iheval}, Control Illusion \citep{controlillusion}, and Many-Tier Instruction Hierarchy \citep{manyih} evaluate how faithfully models respect this ordering. Proposed remedies operate at different levels: the representation \citep{ise}, decoding \citep{vsteer,ihdec}, and the training objective \citep{verih}. These methods assume that the hierarchy stays fixed within a session and that priority is determined by the source. Only a handful of works \citep{ihdec} consider the hierarchy and multi-turn dynamics together. StateIF retains source priority for cross-role conflicts, but adds a temporal dimension and an explicit notion of an instruction that is no longer in force.

\subsection{Externalized State in Reasoning}
Reasoning is a natural tool for complex instructions: RAIF \citep{raif} rewards the reasoning process itself, and Light-IF \citep{lightif} and Thinking LLMs \citep{thinkingllms} train models to think before answering. The evidence is mixed, however. ReasonIF \citep{reasonif}, \citet{scalingreasoning}, and \citet{attentionamnesia} all report cases where more reasoning brings little benefit to, or even hurts, instruction compliance. A separate line of work externalizes intermediate information. Self-Notes \citep{selfnotes} writes intermediate states down as they arise; MemGPT \citep{memgpt}, Think-in-Memory \citep{tim}, and A-MEM \citep{amem} maintain external memories; and RL has been used to learn summary-based context management. \citet{forgetting} show that instruction-based forgetting is fragile, with forgotten content leaking at a rate of 1.00 under elicitation. Process supervision, from process reward models to rewards on the reasoning trace, offers a way to train such intermediate steps directly. What StateIF externalizes is a structured \emph{instruction state} rather than free-form notes or factual memory. The state is learned during the cold start and rewarded directly during RL, together with rewards that keep the reasoning from degenerating into repetition or hallucination.

\section{Method}

\subsection{Problem Formulation}
\label{sec:problem}
Consider a dialogue $D = (u_1, a_1, \ldots, u_T, a_T)$, optionally preceded by a system message $u_0$. Every atomic constraint that appears in $D$ has a period of validity. The response $a_t$ at turn $t$ must satisfy the constraints that are still in force, $A_t$, and must no longer reflect those that have lapsed, $I_t$. The accumulation view implicitly assumes $I_t = \emptyset$ and a monotonically growing $A_t$; we are interested precisely in the case where $A_t$ is not monotone and $I_t$ is not empty. This gives two kinds of errors: \emph{omission}, i.e., violating a constraint in $A_t$, and \emph{residue}, i.e., still reflecting a constraint in $I_t$. Metrics that only measure the satisfaction of $A_t$ cannot see the second kind.

\subsection{Instruction State Machine}
\label{sec:ism}

\paragraph{Instruction objects.}
Each atomic constraint is represented as an instruction object
\begin{equation*}
\begin{aligned}
i = (\ & \mathrm{id},\ \mathrm{content},\ \mathrm{source},\ \mathrm{origin},\\
       & \mathrm{scope},\ \mathrm{status},\ \mathrm{by},\ \mathrm{cond}\ ).
\end{aligned}
\end{equation*}
Here $\mathrm{source} \in \{\textsc{system}, \textsc{user}, \textsc{assistant}\}$, with role priority in that order, and $\mathrm{origin}$ is the turn in which the instruction first appears. The scope is one of \emph{session}, \emph{current turn}, \emph{next $k$ turns}, or \emph{conditional}$(P)$. The status takes one of six values: ACTIVE, SUPERSEDED, REVOKED, EXPIRED, DORMANT, or CONFLICT. The field $\mathrm{by}$ records the instruction that caused the latest transition, and $\mathrm{cond}$ stores the trigger predicate of a conditional constraint.

The state $S_t$ at turn $t$ is the snapshot of all instruction objects, and its two projections instantiate the sets of Section~\ref{sec:problem}:
\begin{align*}
A_t &= \{\, i \mid \mathrm{status}_t(i) = \text{ACTIVE} \,\}, \\
I_t &= \{\, i \mid \mathrm{status}_t(i) \in \mathcal{E} \,\}, \\
\mathcal{E} &= \{\text{SUPERSEDED},\ \text{REVOKED},\ \text{EXPIRED}\}.
\end{align*}
DORMANT instructions, whose condition is not triggered, and CONFLICT instructions belong to neither set. A CONFLICT instruction is the losing side of a cross-level conflict. In our data, 78.6\% of them have a mutually exclusive ACTIVE counterpart, so a response that follows the losing side is already penalized on the ACTIVE side, and counting it again as leakage would penalize the same error twice.

\paragraph{Discourse actions and transitions.}
State transitions are driven by the seven discourse actions in Table~\ref{tab:actions}. Two transitions are easy to confuse. A temporary exception (``just this once'') is not a revocation: the original instruction is marked SUPERSEDED by an exception whose scope is bounded in turns, so that it can come back once the scope ends, whereas \textsc{Revoke} is permanent. Conflicts across sources, for example when a user tries to change a system constraint, are arbitrated by role priority: the higher-priority instruction stays ACTIVE and the lower-priority one is set to CONFLICT.

\begin{table}[t]
\centering
\small
\renewcommand{\arraystretch}{1.15}
\begin{tabularx}{\linewidth}{@{}l>{\raggedright\arraybackslash}X@{}}
\toprule
\textbf{Action} & \textbf{State change} \\
\midrule
\textsc{Add} & new instruction $\rightarrow$ ACTIVE \\
\textsc{Override}($i$) & $i \rightarrow$ SUPERSEDED (by the new value) \\
\textsc{Revoke}($i$) & $i \rightarrow$ REVOKED \\
\textsc{Except}($i$, scope) & $i \rightarrow$ SUPERSEDED by an exception with a bounded scope; when the scope ends, the exception $\rightarrow$ EXPIRED and $i \rightarrow$ ACTIVE \\
\textsc{Conditional}($P$) & new instruction $\rightarrow$ DORMANT; ACTIVE while $P$ holds, DORMANT otherwise \\
\textsc{Reaffirm}($i$) & no change; probes spurious conflict detection \\
\textsc{Expire}($i$) & $i \rightarrow$ EXPIRED \\
\bottomrule
\end{tabularx}
\caption{Discourse actions in ISM and the state transitions they trigger.}
\label{tab:actions}
\end{table}

\paragraph{Why discrete states.}
A revoked instruction and an instruction with very low priority are not interchangeable. Revocation requires that no trace of the constraint remain in the output, whereas low priority only means that the constraint loses when it competes with another one. Scalar priorities, attention weights, and role embeddings cannot express this distinction, which is why ISM keeps the status as a categorical variable.

\paragraph{Conflict taxonomy.}
We categorize conflicts by their semantics rather than from an attacker's perspective, which yields 12 types: same-type satisfiable (C1), same-type mutually exclusive (C2), orthogonal across types (C3), explicit override (C4), implicit override (C5), revocation (C6), exception (C7), conditional activation (C8), benign cross-level conflict (C9), malicious cross-level conflict (C10), requests after privilege downgrade (C11), and unsatisfiable constraints (C12). Each type is aligned with a primary discourse action, a correct behavior, and a typical failure, such as a compromise output under C2 or residual traces of a revoked constraint under C6.

\subsection{Simulating Instruction Dynamics}
\label{sec:data}
Figure~\ref{fig:overview} summarizes the simulation.

\paragraph{Atomic constraints and verifiers.}
We sample constraint categories by weight from a hierarchical taxonomy of instruction constraints, whose top level covers industry, subject, format, content, and style. An LLM then generates $k$ mutually consistent atomic constraints together with a matching domain query, and composes them into a natural complex prompt. Half of the prompts carry a role description and are placed in the system message; the other half are placed in the user message. Each atomic constraint is paired with a verifier. Constraints that can be checked programmatically receive rule-based code, and the rest receive an LLM rubric scored on five levels (0, 0.25, 0.5, 0.75, 1). Constraints introduced later in the dialogue get their verifiers generated on demand. The same verifiers are used for data selection and for the outcome reward in RL (Section~\ref{sec:outcome}).

\paragraph{Three agents.}
The \emph{user agent} picks the conflict type to probe at each turn according to a quota over the 12 types. The two non-conflicting types, C1 and C3, take 20\% each, so that the model also sees many turns in which nothing should change. It then writes the next user message conditioned on the history, and with probability 0.6 applies up to two linguistic perturbation axes: directness (A1), nested negation (A2), reference distance (A3), burial depth (A4), authority framing (A5), code-mixing (A6), and redundant restatement (A7). The \emph{instruction agent} identifies the discourse actions triggered by the user message, updates the state machine, and outputs a DELTA with its rationale. It then runs a self-check covering coexisting contradictions, misclassified actions (e.g., a temporary exception recorded as a revocation), conditional constraints, unhandled expiration, and cross-level conflicts, and retries whenever the resulting state is invalid. The \emph{assistant agent} produces the response and reasoning trace that serve as the teacher trajectory. Because that trace becomes the supervision signal, whatever the agent is told about the state is what the student will learn to say, so we do not hand it the state table. Every entry is instead re-rendered as a single first-person sentence that gives the constraint verbatim, the turn it came from, how far it reaches, whether it applies this turn and, when it does not, the grounds (Figure~\ref{fig:hint}). Status labels, instruction ids and scope values are withheld: none of them occurs in the dialogue, so a student could not re-derive them. In an earlier version that passed them through, 35.4\% of traces copied a status word and about half referred to the table as an external document. The hint exists only at generation time; neither the training inputs nor the test inputs contain it. Each dialogue runs for 30 turns, and every turn becomes a separate record together with its full state snapshot.

\begin{figure}[t]
\centering
\begin{tcolorbox}[colback=c1,colframe=black!40,boxrule=0.4pt,arc=1mm,
                  left=2mm,right=2mm,top=1.2mm,bottom=1.2mm]
\footnotesize
\setlength{\parskip}{2.5pt}\setlength{\parindent}{0pt}
The system setup has a rule, ``never quote internal prices''. It has the highest priority and a user cannot override or revoke it, so this turn I keep following it.

In turn 1 the user asked for ``reply in Chinese'', but the new requirement in turn 3 replaced it, so that one no longer stands and I do not follow it this turn.

In turn 1 the user asked for ``present results as a table''. It holds throughout, so this turn I keep following it.

In turn 2 the user asked me to ``add a risk note whenever a price comes up'', but in turn 5 the user explicitly revoked it, so this turn I do not do it, and I leave no trace of it in the reply.

In turn 3 the user asked for ``reply in English from now on''. It holds throughout, so this turn I keep following it.

In turn 4 the user asked me to ``skip the table this once and write a paragraph'', but that was meant for that one reply and that turn has passed, so this turn I no longer follow it.

In turn 6 the user asked me to ``give the internal prices directly'', but it collides with a higher-priority requirement, so by the rules it yields and I do not carry it out.
\end{tcolorbox}
\caption{The hint given to the assistant agent at T6 of Figure~\ref{fig:teaser}, one sentence per instruction. Each sentence
recovers where the instruction came from, how far it reaches, whether it applies now and, when it does not, why. No status
label, instruction id or scope value appears, because none of them occurs in the dialogue.}
\label{fig:hint}
\end{figure}

\paragraph{State-aware scoring.}
Responses are scored against the state machine of the current turn rather than against the constraints of the first turn. On the ACTIVE side, we verify every constraint in $A_t$ and take the mean as $\bar{s}_A$. On the inactive side, a constraint in $I_t$ that is still followed counts as leakage. For this to be a reliable signal, following an inactive constraint must actually indicate a faulty response, so we apply two filters first. \emph{Polarity filtering} removes inactive constraints whose satisfaction cannot be observed. Once a constraint such as ``do not use the first person'' is revoked, the absence of first-person pronouns no longer tells us whether the model is still obeying it or simply had no reason to use them. An LLM classifies each constraint as requiring presence, requiring absence, or mixed, and only constraints with an observable positive component enter the leakage computation. We use an LLM rather than lexical rules because polarity is semantic: 60.2\% of the constraints that contain a negation word still have an observable positive core. Pure absence constraints make up 37.8\% of all inactive entries. \emph{Compatible-override exemption} handles overrides that leave the old value satisfiable. If ``at most 500 words'' is replaced by ``at most 800 words'', a 400-word response satisfies both. When an inactive constraint and the ACTIVE constraint that replaced it both receive a full score on the same response, we do not count a leak. We require a full score rather than 0.75 here, preferring to miss an exemption over excusing a real leak.

Denoting the filtered set by $I'_t$, the turn-level score is
\begin{equation}
\mathrm{score} = \bar{s}_A - \eta \cdot \frac{\bigl|\{\, j \in I'_t : s_j \geq 0.75 \,\}\bigr|}{|I'_t|},
\label{eq:score}
\end{equation}
where $s_j$ is the verifier output for constraint $j$ and $\eta = 0.2$. A response with $\mathrm{score} > \tau = 0.95$ is considered acceptable, and the same score is used for both gating and ranking. A judgment counts as a leak only at level 0.75 or above, because rubric judges make most of their mistakes at the 0.25 and 0.5 levels, while genuine leaks, where the response fully carries out a revoked requirement, fall at 0.75 and 1. Without this threshold, judge noise alone rejects 77.8\% of turns that have 18 inactive entries; with it, the rate drops to zero. We aggregate leaks as a rate rather than taking the maximum, which maps one leak and three leaks to the same value and leaves 70.9\% of BoN candidates tied. We set $\eta$ well below 1 so that a single leak does not wipe out the ACTIVE score; otherwise, among turns where no candidate passes, ranking would favor a response that satisfies little but leaks nothing. For each turn we sample $N = 5$ candidates and keep the best acceptable one, breaking ties in favor of the longer reasoning trace.

\paragraph{Quality filtering.}
Because the assistant agent sees the state hint, its outputs may rely on information that will not be available at inference time. We discard samples whose response or reasoning copies symbols from the state table (such as instruction ids), refers to the hint as a document, or shows awareness that the hint was injected (e.g., ``the system says\ldots''). We also discard samples with degenerate reasoning (fewer than 200 tokens), looping reasoning (a think-to-body character ratio above 7.7), or repetitive text. Rubric judges only ask whether each constraint is satisfied, so a response that repeats one sentence ten times can still satisfy every constraint. The same detectors are reused as process rewards in RL (Section~\ref{sec:process}).

\subsection{State-Anchored Cold Start}
\label{sec:sft}

Before answering at each turn, the model first writes the current instruction state at the very beginning of \texttt{<think>}, and only then continues with planning, the response, and a final check (Figure~\ref{fig:overview}, bottom). The state block does not make long-range attention stronger. It shortens the retrieval distance instead: the constraints that still matter are copied next to the generation position, and the ones that have lapsed are marked as such, so the model does not have to infer their validity from a long context.

The state block is rendered from the instruction agent's snapshot at that turn, while the reasoning and the response come from the teacher trajectory selected by BoN. CONFLICT entries are not rendered, since the ACTIVE side already carries the winning instruction. Each turn is a separate sample, and only the final assistant turn contributes to the loss. Earlier assistant turns appear in the history as plain responses without reasoning, so the model has to rebuild the full state from the dialogue at every turn instead of copying the previous block. We train with the standard next-token negative log-likelihood over the concatenation of the state block, the reasoning, and the response.

\subsection{RL with Outcome and Process Rewards}
\label{sec:rl}
The reward has two parts. The outcome reward looks only at the final response, and the process rewards look only at the reasoning.

\subsubsection{Outcome Reward}
\label{sec:outcome}
Every instruction is checked at every turn, and each check is routed by verifier type: rule-based code returns a value in $\{0, 1\}$, and an LLM rubric returns one of five levels. Let $v_i(a)$ denote the result for constraint $i$ on response $a$. The current state decides the sign and weight of each check:
\begin{equation}
\begin{split}
R_{\mathrm{out}}(a) ={}& \frac{\sum_{i \in A_t} w_i\, v_i(a)}{\sum_{i \in A_t} w_i} \\
& - \lambda \cdot \frac{\sum_{j \in I'_t} w_j\, \mathbbm{1}[v_j(a) \geq 0.75]}{\sum_{j \in I'_t} w_j}.
\end{split}
\end{equation}
Hard constraints such as format, length, keywords, and language go to code, which is deterministic and unaffected by judge bias; soft constraints such as style, tone, and content go to the LLM rubric. Without the penalty term, a response that keeps following a revoked requirement can still receive full credit, and turns that revoke or override constraints lose their discriminative power. The weights $w$ are set by the state trajectory rather than tuned by hand: instructions touched by the current DELTA receive larger weights, the targets of \textsc{Revoke} receive a larger penalty weight, and once an \textsc{Except} scope ends, the restored constraint is up-weighted on the turn it comes back. The verifiers used for training and for evaluation are implemented independently, so that rewards and metrics do not share the same errors.

\subsubsection{Process Rewards}
\label{sec:process}

\paragraph{State reward.}
We parse the state block from the policy's reasoning and compare it with the reference snapshot. Instruction ids are internal to the state machine and never appear in the dialogue, so the ids written by the policy need not match the reference ones. We first align predicted entries with reference entries by content, and then compute $R_{\mathrm{state}} = \alpha \cdot \mathrm{F1}$ over (instruction, status) pairs. F1 rather than recall keeps the policy from listing every constraint it can find, and a missing or unparsable block receives zero.

\paragraph{Anti-repetition reward.}
Long reasoning traces tend to loop locally or restate whole passages during RL, and the outcome reward is blind to this: a trace that repeats itself ten times can still lead to a perfect response. We reuse the degeneracy detectors from data filtering and turn them into a graded penalty,
\begin{equation}
\begin{split}
R_{\mathrm{rep}} = -\beta \cdot \max\bigl(\ & \phi(\rho_{\mathrm{win}};\rho_0,\rho_{\max}),\\
& \phi(r_{\mathrm{tb}};r_0,r_{\max})\ \bigr),
\end{split}
\end{equation}
where $\phi(x; x_0, x_1) = \mathrm{clip}\bigl((x - x_0)/(x_1 - x_0), 0, 1\bigr)$, $\rho_{\mathrm{win}}$ is the compression ratio of the worst sliding window of the reasoning, and $r_{\mathrm{tb}}$ is the think-to-body character ratio. The thresholds $\rho_0$ and $r_0$ are set at a high percentile of the cold-start data, and $r_{\max} = 7.7$ matches the data filter. A truncated trace, i.e., one without a closing \texttt{</think>}, receives $-\beta$ directly.

\begin{figure*}[htbp]
  \centering
    \includegraphics[width=0.48\linewidth]{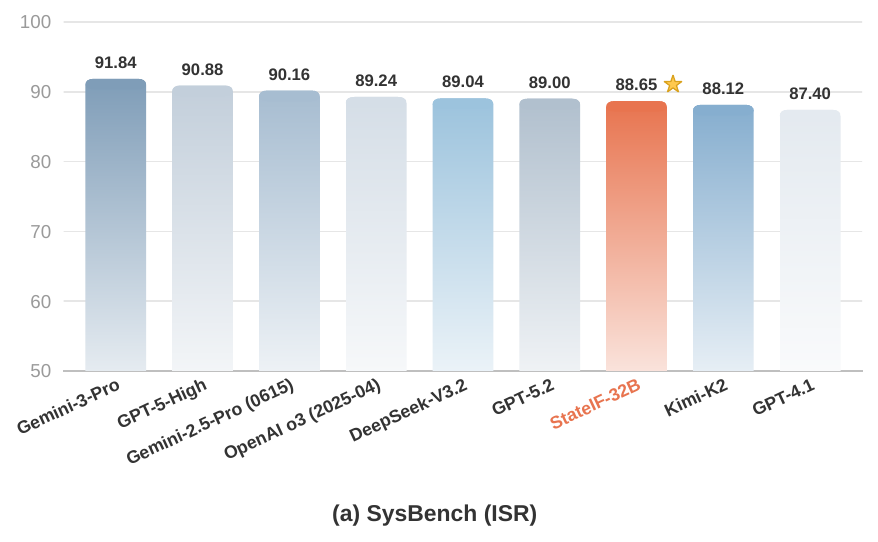} \hfill
    \includegraphics[width=0.48\linewidth]{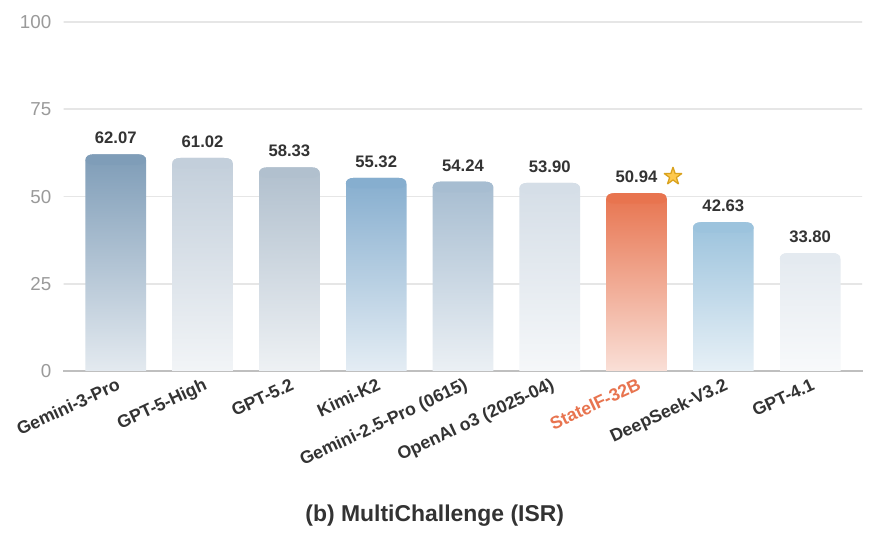} \\
    \includegraphics[width=0.48\linewidth]{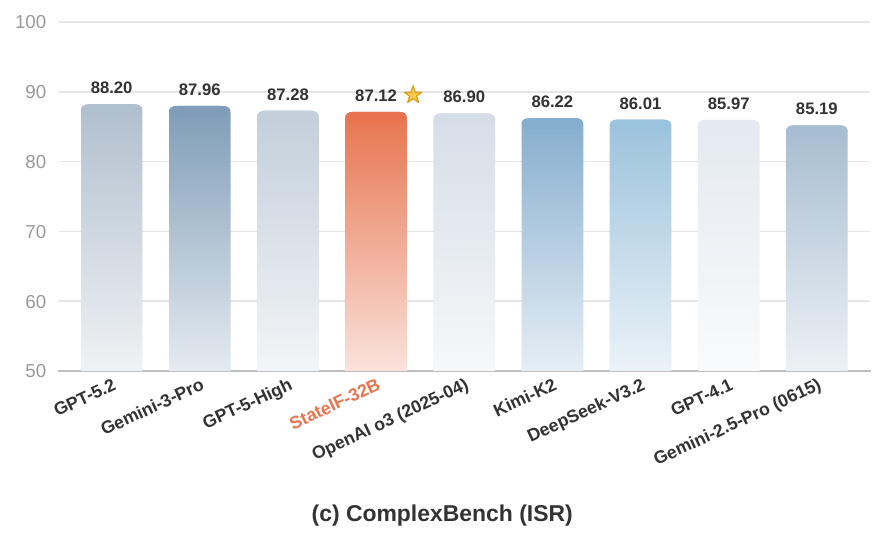} \hfill
    \includegraphics[width=0.48\linewidth]{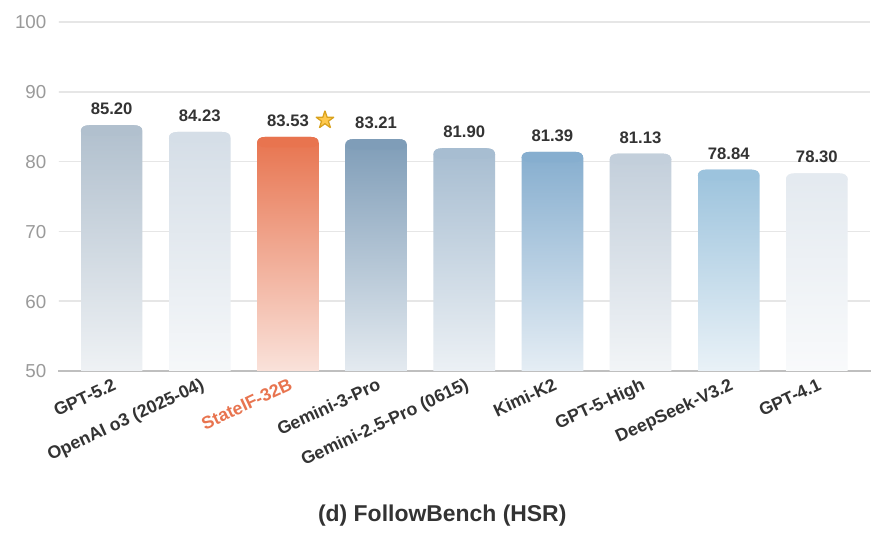}
  \caption{StateIF-32B against frontier proprietary models on the four public benchmarks. All runs have reasoning enabled.}
  \label{fig:frontier}
\end{figure*}
\paragraph{Anti-hallucination reward.}
By hallucination we mean claims about the instruction state that have no support in the dialogue. We distinguish three kinds: \emph{fabricated constraints}, i.e., entries in the state block or the reasoning that were never issued and cannot be aligned with any reference entry; \emph{fabricated transitions}, i.e., claims that a constraint was overridden or revoked when no utterance did so, or references to turns that do not exist; and \emph{fabricated inputs}, i.e., references to material that is not present at inference time (``according to the state table'') or to internal symbols such as instruction ids. The last kind stems from the generation-time hint and can survive the cold start. The first two kinds are detected from the alignment result together with one LLM check that judges, given the dialogue, whether each state claim in the reasoning is supported; the third reuses the pattern detectors from data filtering. The reward is
\begin{equation}
R_{\mathrm{hal}} = -\gamma \cdot \min\Bigl(1,\ \frac{n_{\mathrm{hal}}}{n_{\mathrm{claim}}} + \mathbbm{1}_{\mathrm{inp}}\Bigr),
\end{equation}
where $n_{\mathrm{claim}}$ is the number of state claims in the reasoning, $n_{\mathrm{hal}}$ the number of unsupported ones, and $\mathbbm{1}_{\mathrm{inp}}$ indicates a fabricated input. The two process rewards on the state play different roles: $R_{\mathrm{state}}$ asks whether the model got the status of real instructions right, and $R_{\mathrm{hal}}$ asks whether it invented instructions, transitions, or inputs that do not exist.

\subsubsection{Combination and Optimization}
The final reward is
\begin{equation}
\begin{split}
R &= R_{\mathrm{out}} + R_{\mathrm{proc}}, \\
R_{\mathrm{proc}} &= R_{\mathrm{state}} + R_{\mathrm{rep}} + R_{\mathrm{hal}},
\end{split}
\end{equation}
with $\alpha$, $\beta$, and $\gamma$ chosen so that the range of $R_{\mathrm{proc}}$ stays below that of $R_{\mathrm{out}}$ and the process signal cannot outweigh the task itself. We optimize with GRPO, sampling a group of $G$ responses per prompt and normalizing rewards within the group.

\section{Experiments}

\subsection{Setup}
\paragraph{Data.}
The simulation of Section~\ref{sec:data} produces 17.8K turn-level records. State-aware scoring and the quality filters leave 12.9K of them; the rejections are dominated by samples that exceed the context budget and by traces that name an internal instruction id.

\paragraph{Models.}
Our main experiments use Qwen3-4B, 8B, and 32B \citep{qwen3}. The series still has clear headroom on existing multi-turn benchmarks, which makes the effect of the method itself easier to observe. We compare with ImpRIF \citep{imprif} on the same base model, and place StateIF-32B next to frontier proprietary models as a reference point. The cold start is implemented with LLaMA-Factory \citep{llamafactory} and GRPO with verl \citep{verl}.

\paragraph{Benchmarks and metrics.}
We evaluate on SEQUOR \citep{sequor}, whose conversations run up to 50 turns and whose constraints are added and replaced along the way, and on SysBench \citep{sysbench}, MultiChallenge \citep{multichallenge}, ComplexBench \citep{complexbench}, and FollowBench \citep{followbench}, each with its original metrics. Three of the five are multi-turn. The strictest of their metrics is SysBench SSR.

\begin{table*}[htbp]
    \centering
    \small
    \begin{threeparttable}
    \setlength{\tabcolsep}{0.5mm}
    \begin{tabular}{l|cc|ccc|c|c|cc}
    \toprule
     \multirow{2}{*}{\textbf{Model}} & \multicolumn{2}{c}{\textbf{SEQUOR}}  & \multicolumn{3}{c}{\textbf{SysBench}} & \textbf{\small{\makecell{MultiChallenge}}} & \textbf{\small{\makecell{ComplexBench}}} & \multicolumn{2}{c}{\textbf{FollowBench}} \\
     & \textbf{CSR}  &  	\textbf{ISR}  &  	\textbf{CSR}  &  	\textbf{ISR}  &  	\textbf{SSR}  &  	\textbf{ISR}  &  	\textbf{ISR}  &  	\textbf{HSR}  &  	\textbf{SSR}  \\  
    \midrule									
\textbf{Qwen3-4B}  & 76.84 & 51.29 &  	77.72  &  	60.20  &  	35.20  &  	32.97  &   	77.04  &  	65.56  &  	74.60  \\  
\rowcolor{colorOurs}	
\textbf{$\text{StateIF-4B}$}  &  	\textbf{81.67}$_{\textcolor{blue}{\uparrow 4.8 }}$ &  	\textbf{57.42}$_{\textcolor{blue}{\uparrow 6.1 }}$ &  	\textbf{82.81}$_{\textcolor{blue}{\uparrow 5.1 }}$ &  	\textbf{69.12}$_{\textcolor{blue}{\uparrow 8.9 }}$ &  	\textbf{45.30}$_{\textcolor{blue}{\uparrow 10.1 }}$ &  	\textbf{37.15}$_{\textcolor{blue}{\uparrow 4.2 }}$ &  	\textbf{80.42}$_{\textcolor{blue}{\uparrow 3.4 }}$ &  	\textbf{73.81}$_{\textcolor{blue}{\uparrow 8.3 }}$ &  	\textbf{81.12}$_{\textcolor{blue}{\uparrow 6.5 }}$ \\  

\midrule																			
\textbf{Qwen3-8B}  & 83.70 & 62.55 &  	81.98  &  	66.52  &  	42.00  &  	36.96   &  	81.37  &  	67.89  &  	77.00  \\  
\rowcolor{colorOurs}	
\textbf{$\text{StateIF-8B}$}  &  	\textbf{89.28}$_{\textcolor{blue}{\uparrow 5.6 }}$ &  	\textbf{71.14}$_{\textcolor{blue}{\uparrow 8.6 }}$ &  	\textbf{90.92}$_{\textcolor{blue}{\uparrow 8.9 }}$ &  	\textbf{82.17}$_{\textcolor{blue}{\uparrow 15.7 }}$ &  	\textbf{63.78}$_{\textcolor{blue}{\uparrow 21.8 }}$ &  	\textbf{43.11}$_{\textcolor{blue}{\uparrow 6.2 }}$ &  	\textbf{84.05}$_{\textcolor{blue}{\uparrow 2.7 }}$ &  	\textbf{82.10}$_{\textcolor{blue}{\uparrow 14.2 }}$ &  	\textbf{86.17}$_{\textcolor{blue}{\uparrow 9.2 }}$ \\  

\midrule																			
\textbf{Qwen3-32B}  & 79.41 & 55.83 &  	86.19  &  	73.68  &  	50.40  &  	44.16  & 81.99  &  	71.58  &  	78.58  \\  
\rowcolor{colorOurs}	
\textbf{$\text{StateIF-32B}$}  &  	\textbf{91.50}$_{\textcolor{blue}{\uparrow 12.1 }}$ &  	\textbf{77.25}$_{\textcolor{blue}{\uparrow 21.4 }}$ &  	\textbf{94.05}$_{\textcolor{blue}{\uparrow 7.9 }}$ &  	\textbf{88.65}$_{\textcolor{blue}{\uparrow 15.0 }}$ &  	\textbf{75.26}$_{\textcolor{blue}{\uparrow 24.9 }}$ &  	\textbf{50.94}$_{\textcolor{blue}{\uparrow 6.8 }}$ &  	\textbf{87.12}$_{\textcolor{blue}{\uparrow 5.1 }}$ &  	\textbf{83.53}$_{\textcolor{blue}{\uparrow 11.9 }}$ &  	\textbf{87.46}$_{\textcolor{blue}{\uparrow 8.9 }}$ \\

    \bottomrule
\end{tabular}
\end{threeparttable}
    \caption{Main results on public benchmarks for the Qwen3 series, each with its original metrics.}
    \label{table:main-result}
\end{table*}

\subsection{Main Results}
Table~\ref{table:main-result} compares each StateIF model with its base model. The average gain over the nine metrics is 6.4 points at 4B, 10.3 at 8B and 12.7 at 32B.

The gains are not spread evenly, and where they concentrate is the point of the table. Averaged over the six multi-turn metrics they are 6.5, 11.1 and 14.7 points, against 6.1, 8.7 and 8.6 on the three single-turn ones. At 4B the two are indistinguishable; the gap opens with scale, and at 32B the multi-turn gain is close to twice the single-turn one. The two largest individual gains are also the two strictest multi-turn metrics: SysBench SSR, which credits a session only for its unbroken prefix of correct turns, rises by 21.8 points at 8B and 24.9 at 32B, and SEQUOR ISR, which requires every constraint of a turn to hold, by 8.6 and 21.4. On the single-turn side ComplexBench moves least of all nine metrics at every size (3.4, 2.7, 5.1), which is what we would expect if the method acts on instruction dynamics rather than on constraint composition within one request.

\subsection{Long-horizon Behavior on SEQUOR}
\begin{figure}[t]
\centering
\includegraphics[width=\linewidth]{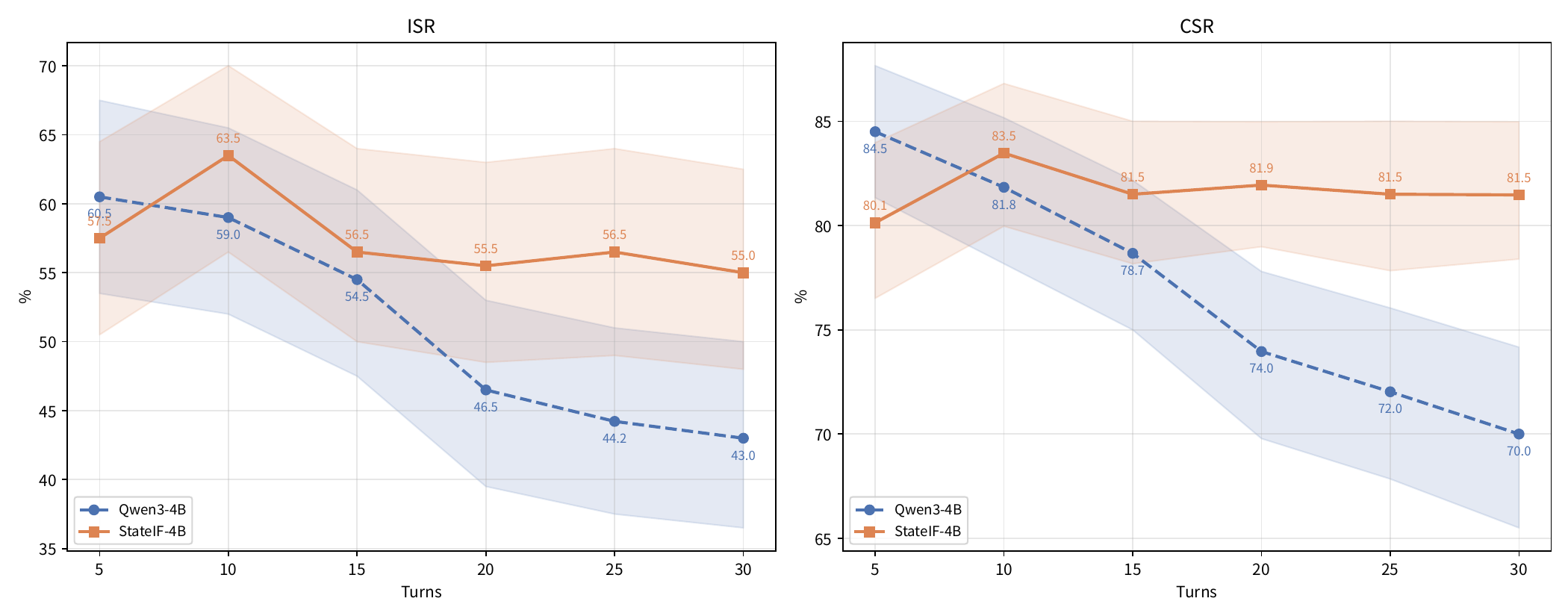}
\includegraphics[width=\linewidth]{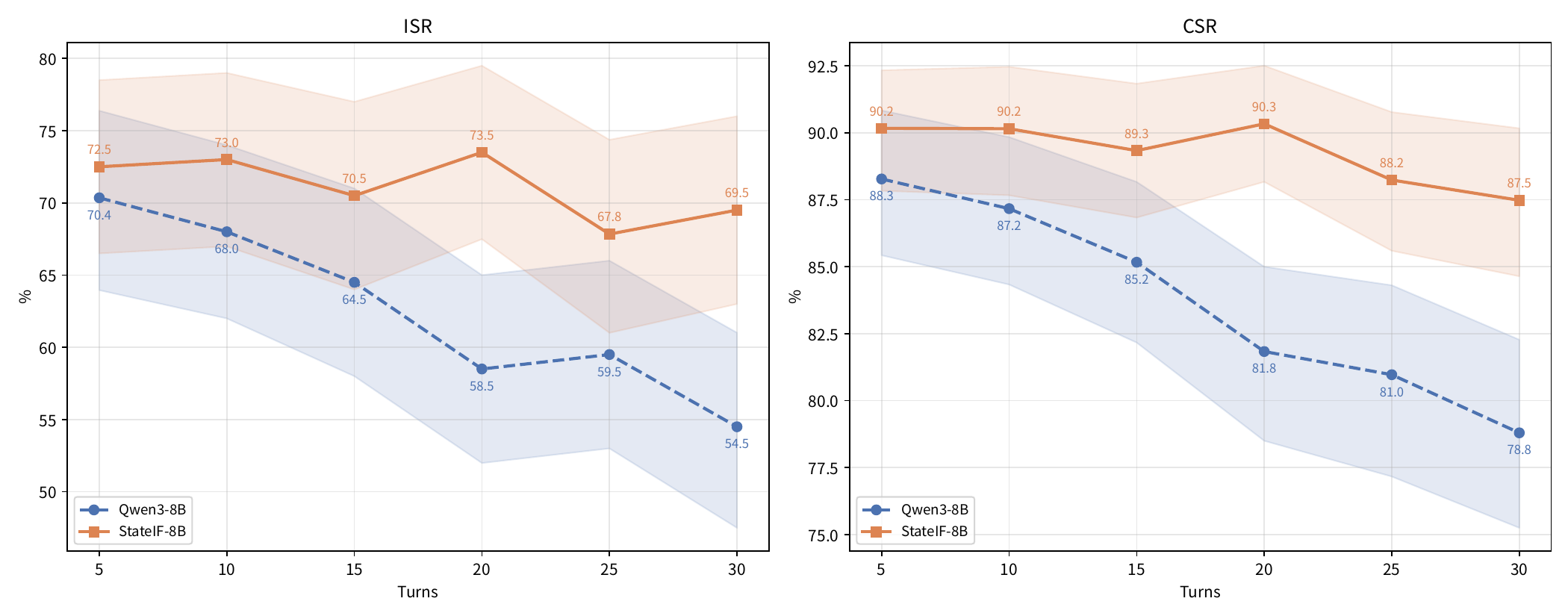}
\includegraphics[width=\linewidth]{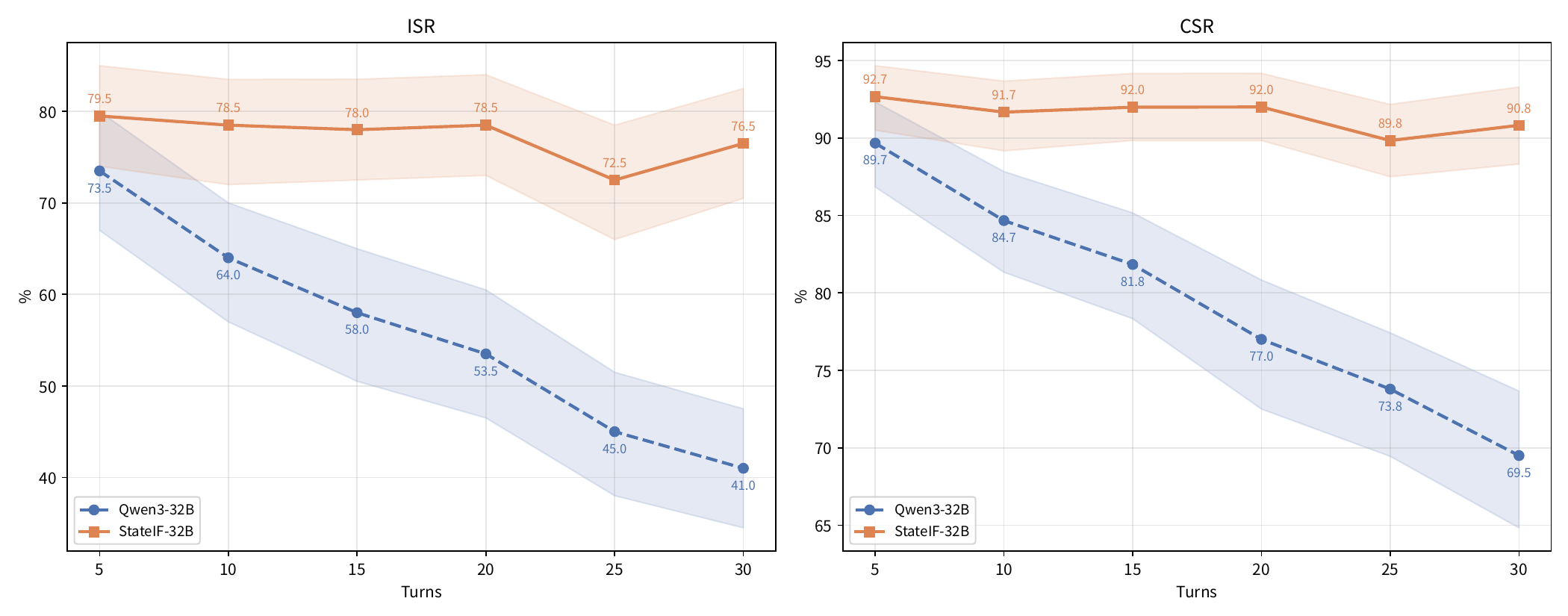}
\caption{Per-turn ISR (left) and CSR (right) on SEQUOR \emph{tuples} over 30 turns, one row per model
size. Dashed lines are the base models, solid lines StateIF; shading is a 95\% bootstrap confidence
interval resampled by conversation. The base curves fall away with turn depth and the StateIF curves
stay close to flat.}
\label{fig:sequor}
\end{figure}

SEQUOR scores a conversation every fifth turn, so each run yields a decay curve rather than a single number, and this is where the effect is clearest (Figure~\ref{fig:sequor}). Over the first 30 turns the base models lose 17.5 (4B), 15.9 (8B) and 32.5 (32B) points of per-turn satisfaction between the first scored turn and the last; StateIF loses 2.5, 3.0 and 3.0. The ranking of the starting points barely changes --- Qwen3-32B in fact opens highest of the three base models, at 73.5 --- so what the training alters is not the model's ability to satisfy a fresh constraint set but its ability to keep doing so as the set is edited. Qwen3-32B is the extreme case: it begins above both smaller models and ends below both of them, which is also why its base SEQUOR scores in Table~\ref{table:main-result} fall below Qwen3-8B's.

\subsection{Comparison with ImpRIF}
ImpRIF \citep{imprif} targets the implicit reasoning structure of single complex instructions, whereas StateIF targets how instructions change over a dialogue, so the two should help in different places. Table~\ref{tab:imprif} compares the two on Qwen3-32B. The split comes out close to that prediction on three of the four benchmarks: StateIF leads on SysBench SSR by 6.1 points and on the two single-turn benchmarks by 1.9 and 5.1. MultiChallenge is the exception and goes the other way by 6.9 points. Its four categories are instruction retention, inference memory, version editing and self-coherence, none of which involves a constraint being withdrawn, so a dialogue there is closer to ImpRIF's accumulating setting than to ours; we read the result as a limit on where the lifecycle view pays off rather than as noise.

\begin{table}[t]
\centering
\footnotesize
\setlength{\tabcolsep}{5pt}
\begin{tabular}{@{}lcc@{}}
\toprule
\textbf{Benchmark} & \textbf{ImpRIF-32B} & \textbf{StateIF-32B} \\
\midrule
SysBench (SSR)       & 69.20 & \textbf{75.26} \\
MultiChallenge (ISR) & \textbf{57.80} & 50.94 \\
ComplexBench (ISR)   & 85.22 & \textbf{87.12} \\
FollowBench (SSR)    & 82.39 & \textbf{87.46} \\
\bottomrule
\end{tabular}
\caption{StateIF against ImpRIF on the same base model, Qwen3-32B.}
\label{tab:imprif}
\end{table}

\subsection{Comparison with Proprietary Models}
Figure~\ref{fig:frontier} places StateIF-32B alongside frontier proprietary models. A 32B open model trained on 12.9K synthetic turns lands inside the frontier band rather than at the bottom of it: on SysBench it is seventh of nine and 3.2 points off the best run, on ComplexBench fourth and 1.1 points off. MultiChallenge is the weakest placement, seventh of nine, consistent with the ImpRIF comparison above. 

\subsection{Analysis}
\label{sec:analysis}

\paragraph{Where in a session the gain appears.}
SysBench scores five turns per session and reports per-turn ISR, which locates the gain within a session rather than across it. At 4B the base model falls 18.2 points from the first turn to the fifth (70.5 to 52.3) and StateIF falls 12.6 (76.8 to 64.2); the two curves start 6.3 points apart and end 11.9 apart. The larger models hold flatter still: StateIF-32B moves 4.2 points across the five turns. The gain therefore comes less from answering the opening turn better than from not degrading afterwards, which is the same pattern the SEQUOR decay curves show over 30 turns.

\section{Conclusion}
We presented StateIF, which starts from the fact that instructions change over a multi-turn dialogue. It uses the Instruction State Machine to formalize constraints as explicit states with a lifecycle, and builds the simulation of instruction dynamics, the cold start, and an RL reward with outcome and process parts on this single formalization. Across the Qwen3 series, the improvement is 6.4 to 12.7 points on average, and the gain is concentrated in the later turns of a session rather than the first: base models lose 16 to 33 points of per-turn satisfaction over 30 SEQUOR turns where StateIF loses 3.

\bibliography{custom}

\end{document}